\documentclass[graybox]{svmult/styles/svmult}

\usepackage{mathptmx}       
\usepackage{helvet}         
\usepackage{courier}        
\usepackage{type1cm}        
\usepackage{makeidx}         
\usepackage{graphicx}        

\usepackage{multicol}        
\usepackage[bottom]{footmisc}
\usepackage{courier}
\usepackage{url}
\usepackage{amsmath}
\usepackage{amsfonts}
\usepackage{hyperref}

\usepackage[textfont=footnotesize]{subfig}

\usepackage{color}
\usepackage{soul}
\definecolor{orange}{rgb}{0.93, 0.53, 0.18}

\makeindex             

\begin{document}

\title*{Tactile sensing}
\author{Lorenzo Natale, Giorgio Cannata}
\institute{Lorenzo Natale, \at Istituto Italiano di Tecnologia, via Morego 30, 16163, Genova, Italy, \email{lorenzo.natale@iit.it}. \\ Giorgio Cannata, \at DIBRIS, Universit\`{a} degli Studi di Genova, Via All'Opera Pia, 13 -- 16145 Genova, \email{giorgio.cannata@unige.it}.}
%
%
\maketitle

This is a post-peer-review, pre-copyedit version of an article published in Humanoid Robotics: A Reference, Springer. The final authenticated version is available online at: \href{https://doi.org/10.1007/978-94-007-7194-9\_110-1}{https://doi.org/10.1007/978-94-007-7194-9\_110-1}
\\
\\
Cite this Chapter as:\\
Natale L., Cannata G. (2017) Tactile Sensing. In: Goswami A., Vadakkepat P. (eds) Humanoid Robotics: A Reference. Springer, Dordrecht. https://doi.org/10.1007/978-94-007-7194-9\_110-1
\\
\\


\abstract{Research on tactile sensing has been progressing at constant pace. In robotics, tactile sensing is typically studied in the context of object grasping and manipulation. In this domain, the development of robust, multi-modal, tactile sensors for robotic hands has supported the study of novel algorithms for in-hand object manipulation, material classification and object perception. In the field of humanoid robotics, research has focused on solving the challenges that allow developing \emph{systems} of tactile sensors that can cover large areas of the robot body, and can integrate different types of transducers to measure pressure at various frequency bands, acceleration and temperature. The availability of such systems has extended the application of tactile sensing to whole-body control, autonomous calibration, self-perception and human-robot interaction.
The goal of this
Chapter is to provide an overview of the technologies for tactile sensing, with particular emphasis on the systems that have been deployed on humanoid robots. We describe the skills that have been implemented with the adoption of these technologies and discuss the main challenges that remain to be addressed. }

\section{Introduction}
\label{sec:introduction}

For many years robotic researchers have looked at vision as the primary source of information to guide robot behavior, while control of interaction forces has been approached mainly with the use of force sensors either at the end-effector or at the joint level.


Operational safety, especially in human-populated environment requires that robot can not only avoid but also detect collisions. Although the former can be done efficiently with visual sensors (especially active sensors, like lasers or infrared) the latter clearly needs to rely on sensors capable of measuring contacts. Force and torque sensors can provide indirect measure of collisions, however, they can do so only if contact happens in certain
parts of the robot body and by relying on an accurate model of the robot.
Safety, especially in presence of humans, needs tactile systems able to cover the whole robot so that unexpected collisions can be detected anywhere.

If industrial robots are programmed to reduce contact with the environment as much as possible, future robotic systems will actually rely on it for proper operation. Researchers are developing algorithms for whole-body manipulation, in which the robot exploits the interaction with the environment to achieve sophisticated behavior (e.g. climbing a stair while holding to the rail, stepping up debris while leaning on supports with the hands, etc).

Human-robot interaction can also greatly benefit of distributed tactile systems. Humans rely on physical contact for communication (like tapping on someone's arm or shoulder to attract his attention, grabbing him to teach how to carry out a task), and likely they will expect a humanoid robot to be able to react appropriately to similar forms of interaction.

Technologies for tactile sensors have been studied exensively in the literature and researchers have proposed many prototypes of sensors which provide accurate response to force or pressure. Implementing tactile systems, however, requires solving additional problems. Tactile sensors are subject by nature to physical stress and need to be reliable and robust against wear and tear. They should have large dynamical range. The hands of a robot, for example, should be equipped with sensors capable of detecting soft touch, when manipulating objects, but also large pressures when supporting the weight of the robot. Similar considerations hold for the frequency response: tactile sensors should detect static contact as well as respond to high temporal and spatial frequency components for detecting slip and discriminate texture. In terms of coverage, the sensor should detect multiple contacts possibly with no \emph{dead spots}, and even be suitable to cover movable parts of the robot (like the joints).

From the mechanical point of view, a favorable property is compliance. It
helps reducing damage either to the robot or the environment by dumping collisions,
and it aids manipulation by increasing contact friction.

Finally, a tactile system should be \emph{affordable}. This implies low cost of manufacturing, deployment and calibration. To fit on a robot a tactile system should use small electronic components for signal conditioning and digital conversion and need a reduced number of wires to route the information from the sensor to the processing units.

For these reasons the development of appropriate tactile systems is an ambitious, technological task which has challenged research for many years.
However, some tactile systems have been proposed in the literature and successfully deployed on humanoid robots, including whole-body systems and compact tactile sensors for humanoid hands. These systems, in turn, have allowed researchers to advance the state-of-the-art in humanoid robot control and cognition. It is today not surprising to see humanoid robots able to operate in contact with the environment, perform delicate object manipulation and object discrimination tasks while relying on pure touch.

The goal of this Chapter is to provide an overview of the recent advancement in robotic touch. In Section~\ref{sec:technologies} we begin with an overview of tactile technologies to give the reader a general understanding of the transduction principles supporting tactile sensing. Section~\ref{sec:tactile-systems} complements this overview with a description of tactile systems that have been successfully deployed on humanoid robots, separated in full body systems and sensors for hands. In Section~\ref{sec:middleware} we discuss the problems of calibration of tactile systems and representation for data processing. In Section~\ref{sec:reactivecontrol} and Section~\ref{sec:perception} we revise applications of tactile feedback respectively for robot control and perception. Finally, in Section\ref{sec:conclusions} we draw the conclusions and discuss the open challenges in tactile sensing.

\section{Technologies}
\label{sec:technologies}

Most of them have been demonstrated at bench-top prototype level, some have been integrated on robots for operational demonstrations, and finally, a few, have been integrated into microcircuits.
There exists a very large number of solutions and technologies proposed for the development of tactile sensors.

A tactile sensor is composed of a supporting element coupled with an electro-mechanical transducer converting a pressure (stress) or a deformation (strain) into a voltage or an electric current. Often, the transducer requires some type of driving voltage or excitation (with the exception of piezoelectric transducers), while the output signal always needs signal conditioning before sampling.

Therefore, the transducer technology is only one of the elements affecting the design of a tactile sensor. In fact, at the system level the development of large scale tactile systems involves significant problems related to their deployment, i.e. \emph{wiring} (routing tactile and driving signal to matrices formed by several -- up to thousands -- taxels) and \emph{embedded electronics} (required to develop self contained tactile systems portable to different robot platforms).

\subsection{Piezo-resistive Sensors}
\begin{figure}[t]
\centering
\includegraphics[scale=0.3]{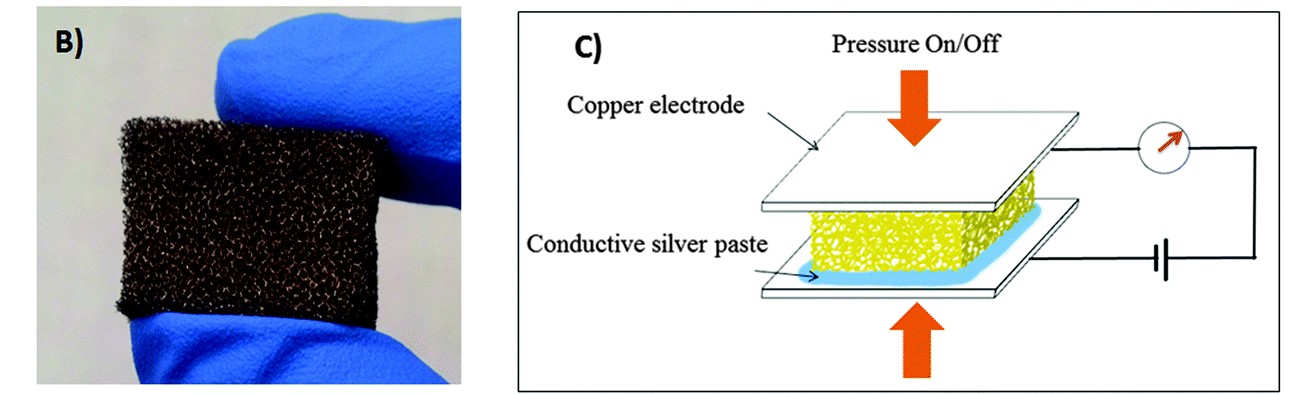}
\caption{Piezo-resistive transducer principle: a piezo-resistive elastomer is sandwiched between electrodes. Source \cite{PiezoResistiveSensor}.}
\label{PRS-figure}
\end{figure}

Piezo-resistive transducers are one the first technologies adopted to develop tactile sensors. The basic design consists of two electrodes (facing each other or interdigitated) bridged by a deformable elastomer loaded with a  conductive filler (e.g. graphite) to reach the so called \emph{percolation threshold}. In this condition the transducer has a high resistivity (typically $>10$~M$\Omega\cdot$cm) in absence of load.  As an external pressure is applied to the elastomer the resistance between the electrodes drops down to resistances of the order of a few K$\Omega$ or less (Figure \ref{PRS-figure}).

These devices require a constant driving voltage. Measurement can be easily performed using a voltage divider formed by the sensor together with a precision reference resistor. Since the resistance variation has quite a large range it is usually possible to exploit the analog-to-digital (A/D) converter input range of the data acquisition system without the need of complex electronic front-end.

Noise can affect piezo-resistive sensors. A first source of noise is intrinsic in the transducer because the effect of the pressure is to change the resistivity of the material by creating conductive pathways. In this respect, special elastomers, e.g. \emph{quantum tunnel composites} exhibit significantly lower noise. The second one, typically adopted in custom made designs, is originated at the electrode-elastomer interface; silver coated electrodes and conductive adhesive can mitigate the problem, but can make miniaturization difficult and prevent manufacturing of large arrays of sensors.

The development of 2-dimensional arrays of piezo-resistive transducers is usually based on the sequential scanning of the rows and the columns: an input multiplexer drive the rows while a de-multiplexer scans the columns. This solution is fairly simple: for a $N\times M$ array only $N+M$ \emph{wires} and a single A/D can be used to acquire all the data. Unfortunately, cross-talk between adjacent taxels affects the measurements~\cite{Hillis1982}. This problem can be solved by grounding the inactive rows and columns during the scanning~\cite{Shimojo2004}, at the cost of increasing significantly the complexity of the driving electronics. An alternative, simpler solution, is to use a hybrid hardware-software compensation methods \cite{CannataMaggiali2006}.

The pressure response is generally \emph{non-linear} and depends on the physical characterization of the polymer and filler which can be estimated using a calibration procedure.

Resistive elastomers are commercially available, and are typically supplied as sheets or laminated foils suitable for planar or cylindrical geometries; more complex geometries (e.g. double curvature surfaces as in the fingertips) may require different manufacturing solutions.

In the past few years a relevant interest emerged for the development of pressure transducers based on \emph{carbon nanotubes} or \textit{graphene}  \cite{Chun2015}, because water based suspensions of carbon nanotubes or graphene can be printed using ink-jet technology. Finally, a very particular usage of piezo-resistive transducers for tactile sensing is based on \emph{electro impedance tomography}~\cite{Kato2007}. In this method no taxels are manufactured, but contacts can be reconstructed using voltage measurements taken at the boundary of the sensitive area.

\subsection{Capacitive Sensors}
\begin{figure}[b]
\centering
\includegraphics[scale=0.5]{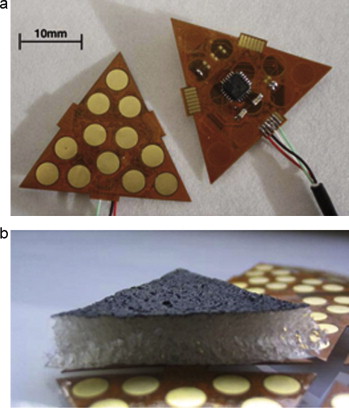}
\caption{Capacitive sensor \cite{CannataMaggiali2006}.}
\label{CS-figure}
\end{figure}

The basic principle of Capacitive sensors consists in two opposite electrodes sustained by an elastic support and separated by a dielectric. This configuration creates a capacitor, whose value can be computed as:

\begin{equation}
C = {\epsilon}_0 {\epsilon}_r \frac{A}{x}
\label{CapacitanceModel}
\end{equation}

where $x$ is the distance between the armatures, $A$ is the area of the electrodes, and ${\epsilon}_0 {\epsilon}_r$ are the \emph{electric constant} and the \emph{relative dielectric constant} (specific of the dielectric material), figure \ref{CS-figure}.
As an external pressure is applied to the elastic substrate the distance $x$ between the electrodes changes increasing the capacitance of the taxel. A complete model of the taxel response can be used for optimizing the sensor design as described in~\cite{LeMaiolino2011}.

Capacitive transducers require an active excitation and demodulation in order to compute the measurement. For this reason, they have received increasing attention in robotics only after their use in touch screens has pushed the development of affordable capacitance-to-digital integrated circuits (CTDs). Time based techniques or frequency based techniques are typically implemented into commercial CTDs which embed fully integrated excitation circuits, electronic front-end and data acquisition in a single chip. However, since most of these devices are designed to detect large capacitance changes (e.g. to implement touch buttons), only a few provide a signal-to-noise ratio (S\//N) suitable for accurate tactile sensing (S\//N $>$ 20~dB).
Another major limitation of existing CTDs is their response time. As the sensor output is computed by averaging sequences of measurements, high S\//N response is typically obtained with output rates significantly larger than 1~ms.

The development of 2-dimensional arrays of capacitive transducers is usually done by driving groups of adjacent taxels with a single CTD (some models provide interface for up to 30 taxels) and by transferring the read-out to a host computer using serial lines or busses. Chip-to-chip data links \cite{schmitz2011} allow to largely reduce the number of lines required to drive sensors with hundreds of taxels.
Flexible \emph{printed circuit boards} (PCBs) \cite{schmitz2011} or semi-rigid PCBs \cite{mittendorfer11} can be used to support the electronics and add additional ground layers to
provide shielding against electrical interference.

At circuit level a technique provided by some CTD manufacturers, called \emph{shielding},  allows compensating for parasitic capacitance (among the circuit wires and with external components) and eliminate cross-talk between taxels.
The response to pressure is generally \emph{non-linear} and a calibration procedure is required to determine numerically the accurate \emph{pressure-to-resistance} relationship, in particular if the sensor is bent over curved surfaces.

Recently, carbon nanotube based conductive inks have been proposed to design capacitive based tactile transducers~\cite{Cagatay2015}. This would make possible to simplify the manufacturing process and to reduce the costs.

\subsection{Piezo-electric Sensors}
Piezoelectric transducers have been used for a long time as tactile and vibration sensors in particular. Compared to other transduction technologies  piezoelectric devices do not respond at low frequencies (zero output in response to steady pressure), but have quite a dynamic response, often larger than 10~KHz. 

\begin{figure}[t]
\centering
\includegraphics[scale=0.3]{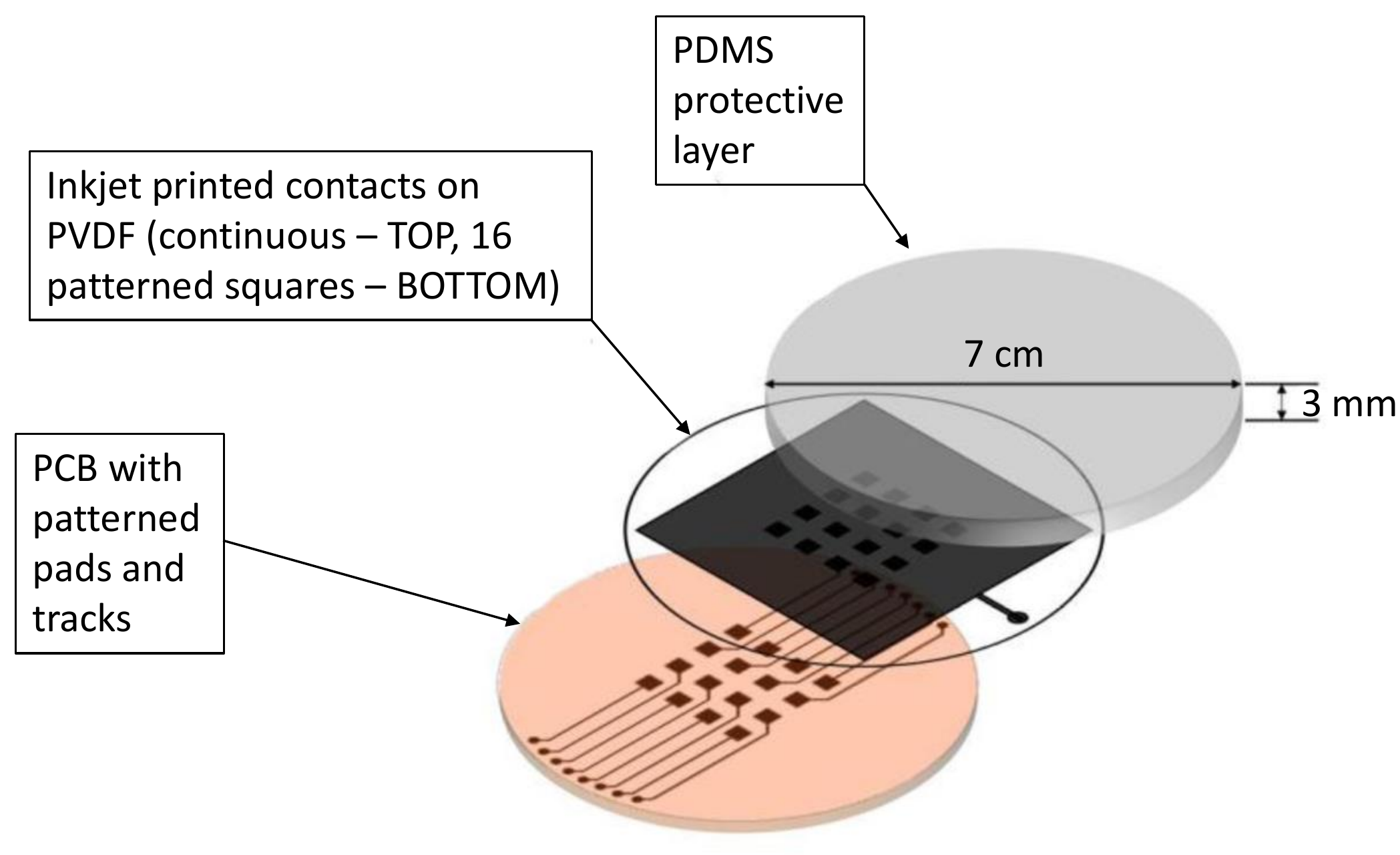}
\caption{Piezoelectric sensor \cite{GastaldoPinna2014}.}
\label{PZS-figure}
\end{figure}

Piezoelectric transducers for the development of tactile sensors are largely based on piezo polymers or ceramics. Among these the \emph{polyvinylidene-fluoride} (PVDF) is perhaps the  most widely used, figure \ref{PZS-figure}.
Piezo materials subject to mechanical stress accumulate electric charges, which are collected by conductive electrodes and amplified to generate a voltage. Therefore, while piezo transducers do not require external electrical excitation, the electronic front-end is complex because it requires high input impedance and amplifiers with low noise. This makes the development of embedded sensor difficult with commercial components.

Since piezo materials are sensitive to different types of mechanical stresses (pressure, bending moments etc.), the design of tactile systems must be done by carefully designing the geometry of the system to capture the desired mechanical input.

Large area tactile systems based on piezoelectric transducers have been proposed. In~\cite{Taichi2006} each taxel is formed by a deformable PVDF foil sandwiched between two polyurethane foam layers.
Finally, integrated arrays have been developed by coupling a pressure sensitive PVDF film on the gate of a FET transistor creating the so-called POSFET device~\cite{Dahiya2014}.

\subsection{Optical Sensors}
\begin{figure}[b]
\centering
\includegraphics[scale=0.3]{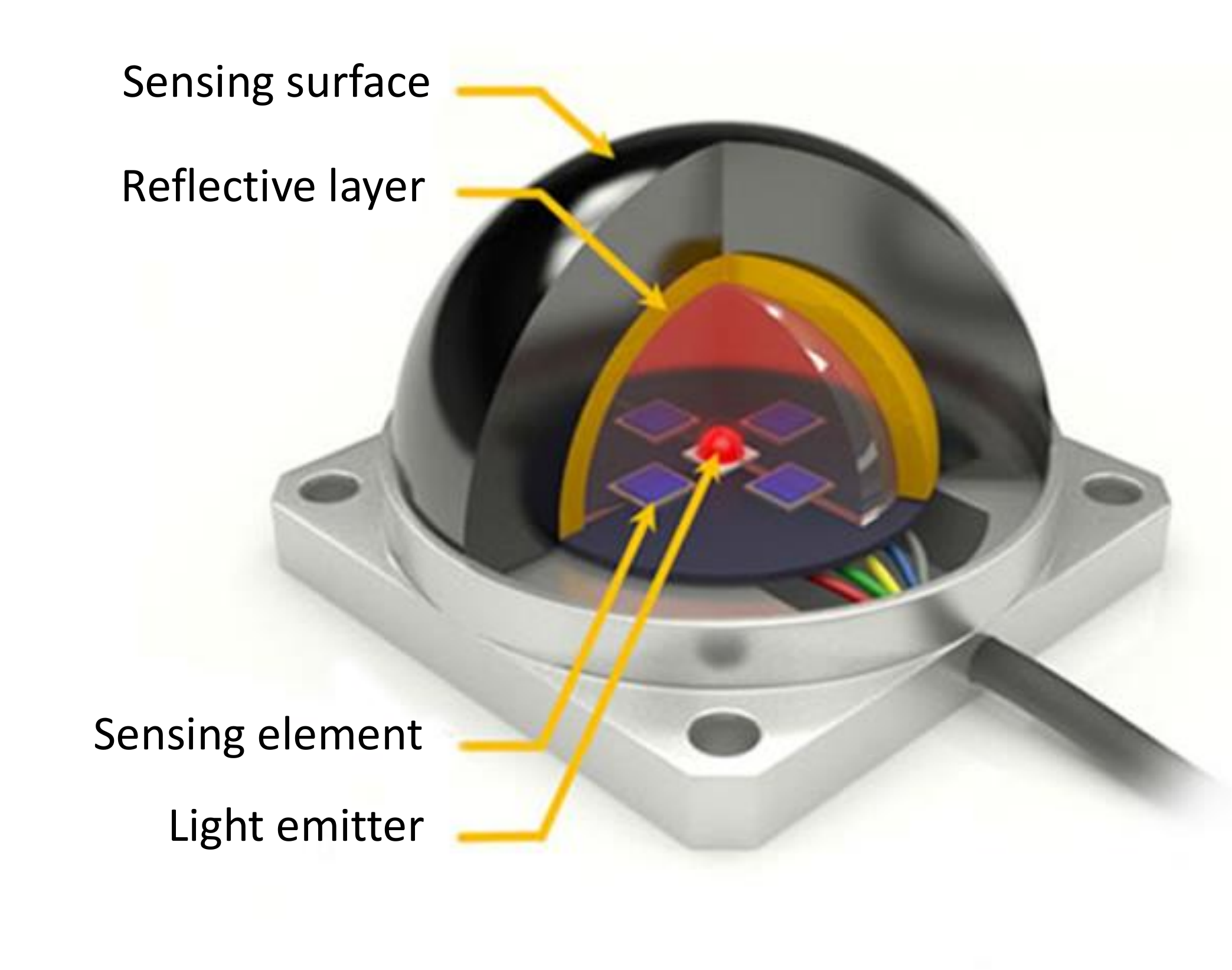}
\caption{Commercial optic sensor (OptoForce Ltd.).}
\label{OS-figure}
\end{figure}

Optical transducers have been used over the years in many different ways to develop tactile sensors.
The very first large area robot skin developed by Lumelsky \cite{CheungLumelsky1988} used optical transducers as proximity sensors, a technique also used in recent designs to detect contact with light or extremely thin objects \cite{mittendorfer11}, \cite{DeneiMaiolino2015}.
Other implementations are based on reflectivity measurements \cite{Ohmura2006}, \cite{CirilloNatale2016}, . In this case a Light Emitter Diode (LED) and a photo detector are coupled and covered by a deformable opaque elastomer. When pressure is applied the elastomer is subject to deformation and light is scattered: the variation of reflected light is captured and processed, figure \ref{OS-figure} (from Optoforce Ltd.\footnote{https://optoforce.com/}).

The driving and front/end electronics is fairly simple and the transducer is robust against electro-magnetic disturbances, this makes it possible to use high sampling rates. The development of 2-dimensional arrays of optical transducers is usually done by driving the taxels with a row-column scanning mechanism~\cite{ohmura06}.  However, in large implementations a critical aspect to consider is the power consumption of the LEDs.

There is no analytic model for characterizing the response of the sensor, therefore, an empirical calibration of the device can be required as in ~\cite{CirilloNatale2016}.

\subsection{Magnetic Sensors}
\begin{figure}[b]
\centering
\includegraphics[scale=0.3]{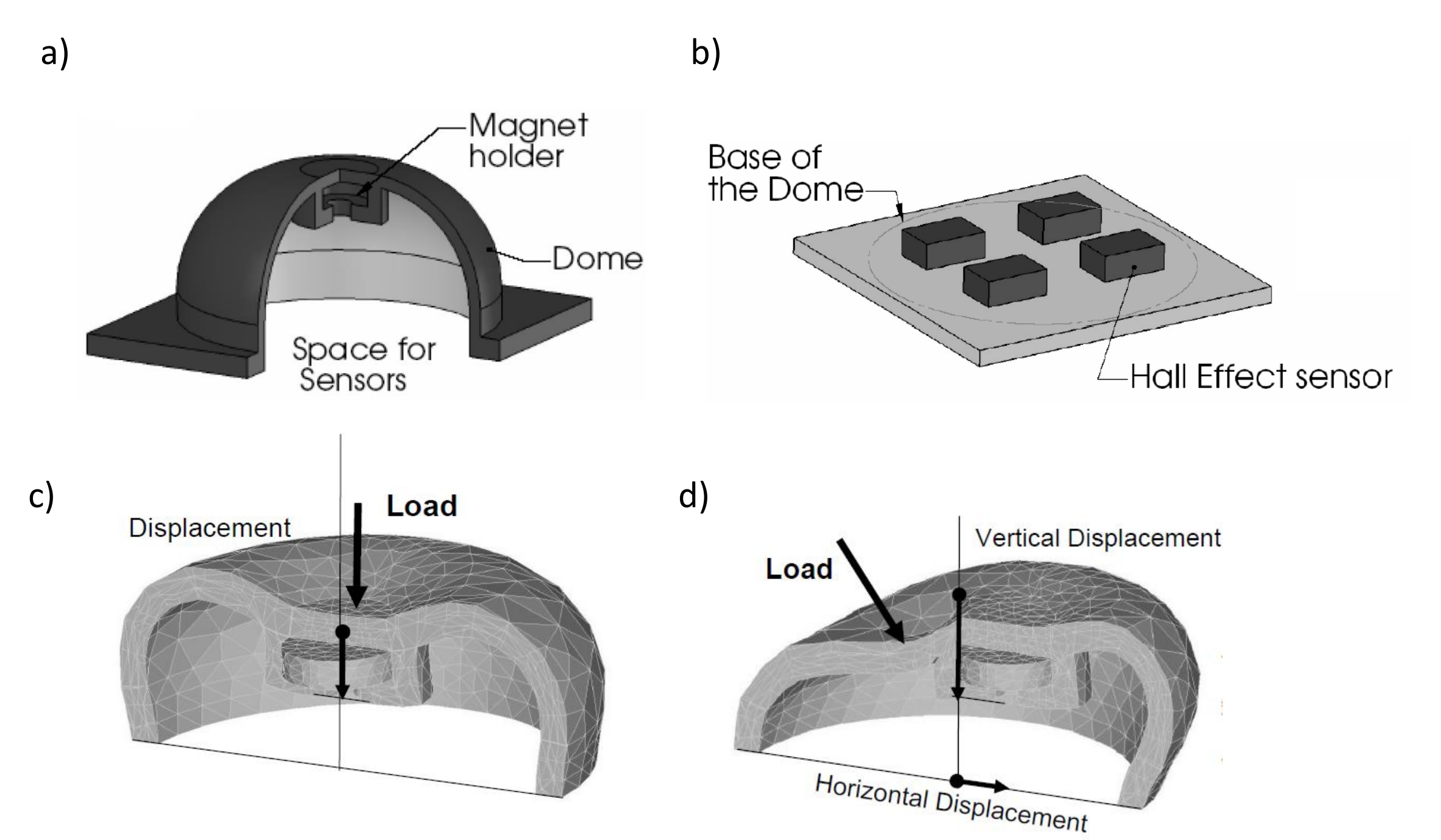}
\caption{Magnetic transducer principle. Image courtesy of Eduardo Torres-Jara \cite{torres-jara06}.}
\label{MS-figure}
\end{figure}

Magnetic transducers, like capacitive transducers, have evolved in the past few years along with the growth of the market of Hall-effect Integrated Circuits (ICs).  The basic design consists of a magnet suspended by an elastic support over the sensitive point of the Hall-effect IC. When external pressure is applied, the relative position and orientation of the magnet changes as well as the measured intensity and orientation of the magnetic field~\cite{torres-jara06,Jamone2015}, figure \ref{MS-figure}.

One of the limitations of these systems is that they could interfere with surrounding ferro-magnetic materials and can be influenced by external magnetic fields. The development of 2-dimensional arrays can be based on the same architectures described for the capacitive based systems.

\section{Tactile Systems}
\label{sec:tactile-systems}
In this Section we describe a few significant tactile systems that have been implemented on humanoid robots, solving system level challenges that made these tactile systems at least in principle portable to other robots. We have considered two major classes of tactile systems: whole body tactile systems (\emph{robot skins}) intended to cover large parts of the robot body, and tactile systems for robotic hands and manipulation.

\subsection{Whole-body systems}

Early attempts to cover large areas of a humanoid robots are~\cite{inaba96,kageyama99,iwata02}. In~\cite{inaba96} the authors proposed a ``sensor suit'' to cover the entire body of a robot. The system provided a binary output (contact versus non-contact), based on resistive principle and it was made of 192 sensing regions on the whole body with variable spatial resolution ranging from 10$\times$5~cm for the legs to 5$\times$5~cm for the arms. The sensitivity of this system was relatively low (4900~KPa), however, it  allowed the implementation of simple touch driven orientation behaviors and whole-body grasping (i.e. caging).


In~\cite{kageyama99} the humanoid robot H4 was equipped with five tactile modules: the chest was covered by a system of 96 sensing points; each upper and fore-arms had 64 sensing and 36 sensing points respectively. Pressure was estimated by measuring change in resistance between electrodes separated by a soft, conductive gel. The measurable pressure was within 0-40~KPa with minimum threshold of 2.5~KPa, with a sampling time of 80~ms.

The first tactile system capable of detecting force has been implemented on the
robot Wendy~\cite{iwata02}.
This system was made of modular units made of a rigid cover
integrating a 6 axis force-torque sensor and a set of \textit{Force Sensing Resistors} (FSR) on its surface. The magnitude, direction and location of the external force was computed by integrating the information provided by the sensors, under the assumption that no moments were generated at the contact point.
A total of 6 units were mounted on the arms and shoulders of the robot Wendy: the system allowed to sample data at 100~Hz and measure 3D force vectors with average standard deviation of 1.2~mm and 1.6~mm for the contact point on the surface and of 0.15~N for the force intensity.

The humanoid robot ARMAR III~\cite{armar} was equipped with artificial skin made of planar skin pads mounted on the front side and back side of each shoulder, and interconnected by CAN-bus links. The sensors pads used resistive technology implemented using a graphite loaded elastomer and electrodes obtained using a flexible PCB~\cite{armar-tactile-sensors}.
%
Data was acquired at the rate of 40~Hz with a 12~bits resolution, in the range of 4\--120~KPa. 

The robot CB$^{2}$~\cite{minato07} was designed to support social interaction with humans and, for this reason, it was fully covered with a soft skin, made of PVDF films covered by silicone rubber. Since PVDF detects the rate of change of the applied stress, this information was integrated to reconstruct the actual contact pressure. Overall the robot was covered by 197 tactile sensors, distributed on the arms, shoulders, torso arms and head, sampled at the rate of 100~Hz.

The robot Kotaro~\cite{mizuuchi06} mounted two types of tactile sensors. The first was
a flexible ``band'' obtained as a sandwich of a force-sensitive conductive rubber and two flexible PCBs. The ``bands'' were wrapped around the robot links and represented one of the first examples of a modular \textit{robot skin} system.
This system had 64 sensing units which could be read individually. The second type of sensors used conductive rubber foam in complex 3D shapes: pressure was estimated by measuring the change of resistance between electrodes.

A highly modular system based on optical technology was described in~\cite{ohmura06}.
The system consisted in modules of 32 sensors which could be adapted to cover non-flat surfaces. Modular units could be connected in a LAN to achieve the maximum number of 65536 sensors. The operational range of the sensing units was in the range of 0\--500~KPa. In~\cite{ohmura07} this system was used to cover a humanoid robot with 1864 sensors; this system was used to demonstrate lifting of a 32~Kg box with whole-body contact.

Semiconductor pressure sensors were employed in the robot RI-MAN~\cite{mukai08}, with the aim of supporting object manipulation and human-robot interaction. The authors used piezoresistive semiconductor pressure sensors which had the diameter of 5.8~mm and could detect the absolute pressure between 40~KPa and 440~KPa. These sensors were mounted on a \textit{comb-like} flexible PCB, which could bend and conform to non-flat surfaces. Tactile sensors on the RI-MAN were located in five places -- namely the chest, the arms and forearms -- for a total of 320 sensing elements. The tactile system was made by elements of 8$\times$8 sensing elements which were refreshed every 15~ms. The range of measured force varied between 0 to 8~Kg over an area of 25$\times$25~mm\textsuperscript{2} (corresponding to a maximum pressure of 126~KPa). The tactile system was employed to measure the force exerted by the robot while lifting a (dummy) human body.

\begin{figure}[b]
\centering
\includegraphics[scale=0.6]{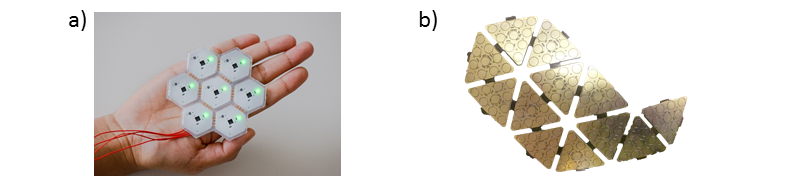}
\caption{Examples of skin systems. Left: the HEX-O\_SKIN (image courtesy of A. Heddergott Munich Institute of Technology). Right: the ROBOSKIN (image courtesy of Istituto Italiano di Tecnologia).}
\label{tactile-systems}
\end{figure}

A multi-modal modular tactile system (called HEX-O\_SKIN) for whole-body sensing was presented in~\cite{mittendorfer11} (Figure~\ref{tactile-systems}). Each module embeds sensors for temperature, 3-axis acceleration and proximity to emulate the human senses of temperature, vibration and light touch. Modules have hexagonal shape, and can be interconnected to form a mesh. Although individual modules are rigid the system can bend at the intersection and conform to curved shapes. A local controller on each module processes data from the sensors at the frequency of 1~Khz and routes it to other modules. This configuration greatly reduces the number of wires, with increased robustness and scalability. A system with 8 modules was initially mounted on a Kuka lightweight robotic arm~\cite{mittendorfer11}.
The HEX-O\_SKIN patches have been recently enhanced to include three capacitive sensors measuring normal force up to 10~N, and Gigabit interface to read the data from a network of cells.  A system of 74 modules has been deployed to cover the upper body of the robot HRP2
\cite{mittendorfer15} and used to implement various tactile behaviors, such as kinesthetic teaching using contact or proximity and whole-body grasping of unknown objects.

The \textit{ROBOSKIN} tactile system~\cite{schmitz10} uses capacitive technology (Figure~\ref{tactile-systems}). In this case sensing units are capacitors obtained by layering a flexible PCB, a deformable dielectric, and conductive Lycra. Pressure deforms the dielectric and varies the distance between the conductive plates, resulting in a variation of capacitance that can be measured by commercial components. Similarly to the HEX-O\_SKIN modular elements are interconnected by a bus. In this case the modules are triangles, which embed 10 pressure sensors and two thermal sensors each: up to 16 triangles can be read at the frequency of 20~Hz by a processing units which broadcasts tactile values on a CAN interface. The \textit{ROBOSKIN} tactile system has been used to cover various robots. The iCub robot~\cite{metta10} mounts a total of 4488 sensors distributed on the hands (including the fingertips, the arms, the torso, the legs and feet-soles). The latest version of the tactile system achieves low hysteresis and good mechanical robustness using a sandwich of three fabrics which form the dielectric, the conductive Lycra and a protective layer~\cite{maiolino2013}. It can detect the minimal pressure of 2\--3~KPa, and the maximum value of 180~KPa. Experiments with the iCub have addressed control of interaction forces on the whole-body~\cite{delprete12,fumagalli12} and visuo-tactile calibration~\cite{roncone15}. Other robots that have been covered with the \textit{ROBOSKIN} system are KASPAR (816 tactile elements) and NAO (324 tactile elements).

\subsection{Tactile sensors for antropomorphic hands}
\label{sec:hands}

In the area of grasping and manipulation a large number of solutions have been investigated to provide hands with tactile sensorization. One of the early example is the Gifu hand II, which is equipped with 6-axis force sensors on the fingertips and a distributed system of tactile sensors based on resistive technology with 624 points on the surface.

A sensorized glove for the NASA Robonaut hand was presented in~\cite{martin04}. The basis for this technology was Quantum Tunneling Composite (QTC), a material which changes resistance with the applied pressure. QTC can be produced in sheets that conform to curved surfaces and are sensitive to forces from a fraction of a Newton to 10~N. The glove provided 33 sensor elements and increased friction and sensitivity by incorporating plastic beads that acted as force concentrators.

The hands of the robot Obrero was equipped with 80 tactile sensors~\cite{torres-jara06}. Each sensor unit had a dome-like shape made of silicone rubber, hosting a small magnet in the inner tip. Four hall-effect sensors at the base of the dome measured the magnetic field: mechanical deformation of the dome modified the magnetic field and allowed estimating the applied pressure. These sensors can detect the minimum force of 0.094~N and were successfully used to grasp unknown objects using tactile feedback alone~\cite{natale06}.

\cite{goger09} described a tactile system for an anthropomorphic hand. This system embedded a PVDF sensor within a resistive pressure sensor. The resistive sensor was made by a set of electrodes covered with a conductive foam: pressure changed the resistance measured between a common, reference, electrode and the individual sensing elements. The PVDF was embedded in the cover of the sensor. The mechanical properties of the cover transmitted vibrations that were picked up by the sensor. This system was used to sensorize the inner parts of the fingers and palm of an anthropomorphic hand. It was used to detect slip, contact points and classify tactile images.
%

\begin{figure}[t]
\centering
\includegraphics[scale=0.5]{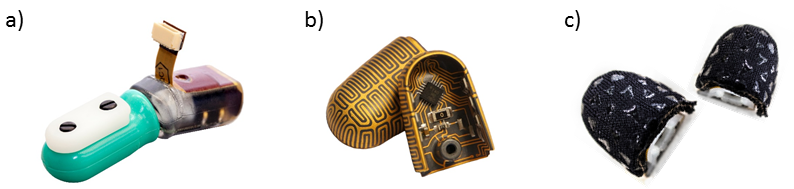}
\caption{Examples of sensorized fingertips. a) The BioTac~\cite{wettels07} (image courtesy of SynTouch, LLC). b) Resistive fingertip realized with Laser-Direct-Structering technique~\cite{koiva13} (image courtesy of Risto Koiva, Bielefeld University). c) The iCub fingertips~\cite{jamali15} (image courtesy of Istituto Italiano di Tecnologia).}
\label{fingertips}
\end{figure}

The iCub hand was equipped with tactile sensors on the fingertips~\cite{jamali15} (Figure~\ref{fingertips}). These sensors were a customization of the iCub tactile technology, adapted to fit on the space available on the fingertips. The resulting system had 12 sensors: it was made by layering a flexible PCB and deformable fabric which provide the dielectric, conductive and protective layers. It can detect the minimal pressure of 3\--4~KPa up to a maximum or approximately 50~KPa.

Laser-Direct-Structuring was used in~\cite{koiva13} to deposit conductive tracks on a 3D structure which formed the fingertip for the Shadow Hand (Figure~\ref{fingertips}). This system used resistive technology and it provided a total of 12 sensing elements with the resolution of about 5.5~mm that can be read by an integrated circuits at the frequency of 1~KHz. This fingertip can sense forces up to 80~N, with a tradeoff between sensitivity and maximum measurable load. It was used in experiments that involve manipulation tasks such as opening and closing jars and folding paper, which are extremely challenging to accomplish without tactile feedback.

Probably the most successful tactile sensor for hands is the BioTac from Syntouch~\cite{wettels07} (Figure~\ref{fingertips}). It is a bioinspired, multimodal fingertip which includes resistive sensors, pressure transducers and temperature sensors. The sensor is made by an elastomer which contains a conductive fluid: deformations of the elastomer produce change of resistance between the electrodes inside the fluid, whereas vibrations are captured by the pressure sensor. The fingertips are therefore sensitive to forces and vibrations. The sensor can detect contacts with the spatial resolution of 2~mm and forces that vary from 0.1~N to 30~N The BioTac has been mounted on many robotic hands: the Shadow Hand, the Barret Hand, the PR2 gripper, the Allegro Hand from SimLab, the JACO Hand, the Hubo hand, the 3Finger Adaptive Gripper from Robotiq and the SDH from Schunk\footnote{http://www.syntouchllc.com/}. The BioTac has been successfully used to solve a large number of tasks (e.g. material discrimination, object recognition, slip detection and re-grasp to improve stability). These tasks are usually solved using Machine Learning because the activation of the sensing elements in response to pressure is difficult to model.

\section{Calibration and Data Representation}
\label{sec:middleware}
Tactile data are generated using different types of transducers, and they are usually placed at discrete points over the robot body. Data are expected to be collected and a properly processed for implementing control and perception tasks. A similar problem is faced when the task at hand requires integration of information from various sensors, attached to different, moving, reference frames (e.g. inertial units or cameras). A possible way to solve these problem is to perform a spatial calibration of the tactile system, so that tactile information can be referred to a reference frame attached to the robot. Cross-calibration with other sensory modalities is another option, especially when the position of the other sensors is known with higher accuracy or when the task requires it.

Another approach is to avoid spatial calibration and work directly on \textit{tactile images} obtained by representing tactile data in two dimensional surfaces. This approach is based on an apparent analogy between tactile data and visual images. Unfortunately, this analogy is only superficial. In fact, geometry and raw data format of transducers for vision consist of 2-dimensional arrays arranged in a rectangular pattern. Specific geometric and optical calibration (pixels' size, image center, lens distortion), can be performed by \textit{state-of-the-art} software (e.g. MATLAB: Computer Vision System Toolbox). Secondly images captured by digital cameras encode a well defined physical quantity: the amount of light captured in a given time interval. Therefore, visual images acquired by physically different cameras (namely devices produced by different manufacturers) are in principle equivalent: it is therefore possible to develop image processing algorithms which are independent from the specific hardware used, and it is possible to define \textit{universal} image features which rely on standard camera models.

Tactile images are neither standard nor based on a standard model.
In a generic skin system taxels are placed over the robot body in order to provide a more or less coarse spatial sampling of the contacts. This has various implications. First, because a robot body can be modeled as a 2-dimensional manifold, taxels are geometrically arranged in patterns that depend on the specific shape of the robot (i.e. different robots, or even different links of the same robot, have different taxels arrangement).  Relative position of taxels belonging to the same robot can even change as the robot moves. Second, taxels could be placed over the robot with a space-varying density and size, to provide higher resolution in selected body areas (e.g. the hands or the fingers), and lower for others (e.g. the torso and the back).
Finally, a robot could mount different types of tactile transducers, in general acquired at different rates, which require distinct representations.

A proper \textit{software abstraction layer} for tactile system can support the development of tactile data processing for control and perception that are unspecific to the actual hardware. This idea, first introduced in~\cite{Hoshino1998} has been recently implemented in a system named \emph{Skinware}~\cite{Youssefi2015b,Youssefi2015a}.

\subsection{Calibration}
\label{sec:calibration}
Some control tasks driven by tactile feedback (see for example Section~\ref{sec:reactivecontrol}) require knowledge of the precise location of taxels with respect to the robot body. In principle the location of the taxels is known in advance when the skin is integrated on the robot; in practice, this knowledge is often inaccurate and affected by uncertainty arising during the deployment of the sensors on the robot surface. Sources of errors are the positioning of the system itself and the deformation of flexile parts when they conform to the surface. The position of taxels, in addition, depends on the robot shape and its kinematics. These observation motivates the need for automatic techniques for the spatial calibration of tactile systems.

In~\cite{delprete11} the authors used information from the F/T sensors on the robot to calibrate the tactile system of the iCub robot. The idea is that, in absence of external torque, it is possible to use the force/torque measures to estimate the point of application of external forces and correlate the estimate with the tactile data. This calibration technique was validated in a real robotic system, and it allowed estimating the position of taxels with the average error of about 7~mm.

An automatic calibration procedure for sensors and actuators is described in~\cite{mittendorfer12}. This technique allowed calibrating the position of the taxels on each HEK-O\_SKIN module by exploiting its rigid structure. The authors used specific motion patterns and the accelerometers embedded in the HEK-O\_SKIN tactile system to determine the structure and parameters of the kinematics chain that describe the robot and the placement of the individual units.

A different approach was described in~\cite{roncone15}, in which the robot performed a visual calibration of the tactile system. The idea in this case was to attach a spatial receptive field to each taxel and to adapt this representation using vision. Adaptation took place by touching the tactile system repetitively in multiple locations using objects. The robot tracked objects using vision and linked their spatial location before contact (as measured by the cameras) with a receptive field anchored to the taxel that got activated after touch. The interesting aspect of this approach is that the tactile system was calibrated with respect to the reference system of the cameras. The authors demonstrated how to use this representation for predicting and avoiding collisions with objects in the visual space.

\begin{figure}[b]
\centering
\includegraphics[scale=0.5]{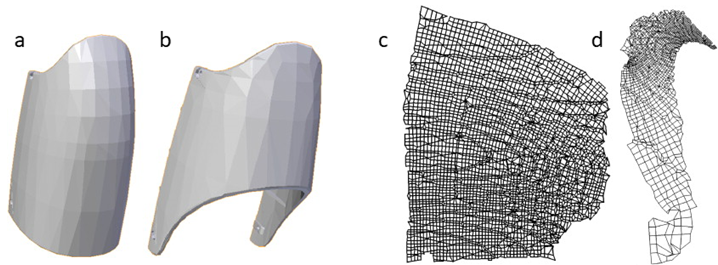}
\caption{Simulated 3-D (a and b) to 2-D (c and d) transformation for 2 parts of the iCub robot for a fine taxel distribution. Source \cite{DeneiCannataMastrogiovanni2015}.}
\label{iCubMaps}
\end{figure}

\subsection{Data Representation}
\label{sec:representation}

Tactile data representation is difficult for various reasons. Taxels are located on curved surfaces without a regular pattern, therefore concepts like \emph{proximity} or \emph{distance}, or operations like spatial filtering cannot be readily applied. In~\cite{CannataDeneiMastrogiovanni2010a} it has been proposed to represent tactile regions (see below) as flat 2-dimensional maps, also called \emph{artificial somatosensory maps} from their analogy with a similar representation found in the cortex of primates. These maps are obtained by \emph{flattening} the robot surface with minimal distortions to preserve the relative location of the taxels.

%
%
%
Once the tactile data are represented as 2-D maps, standard techniques for filtering and processing (e.g. borrowed, from computer vision) can be used. In particular the maps can be re-sampled by software leading to  standard piecewise rectangular grids.
Figure \ref{iCubMaps} shows a simulated 3-D to 2-D transformation of a two parts of the iCub robot.

A further advantage of \emph{somatosensory maps} is that they can be used to calibrate multi-modal systems, layering on a common reference frame data from different sensors. Finally, such maps can be used to plan and control interaction tasks. The advantage is that contact location, force and task errors can be expressed in a bi-dimensional space and conveniently back-projected to the 3-D space when needed (for example to compute robot control actions).

\section{Reactive control}
\label{sec:reactivecontrol}

In this Section we describe applications of tactile systems for control. We separate the discussion in two parts: whole-body control and manipulation.

\subsection{Whole body control}
Whole-body tactile systems have been used to implement various behaviors. The earliest applications involved control of the robot to perform tasks with the whole-body like object caging using the arm and chest~\cite{inaba96,kageyama99} (and more recently~\cite{mittendorfer15}) , or lifting of heavy objects~\cite{ohmura07,mukai08}.

The typical application of tactile sensors is, however, contact avoidance. For such behaviors the main advantage of tactile sensors is that they directly measure or allow to accurately estimate the contact forces. Computing the same information from the torque sensors at the joints is more complicated because it requires computing and removing the torque component that is due to the robot internal dynamics.
An example of this type of behavior can be found in~\cite{kageyama99} where the mobile base of the robot H4 is controlled to retract to reduce the pressure on the chest. Similarly, evasive movements have been implemented using the HEX-O\_SKIN on a Kuka lightweight arm~\cite{mittendorfer11}.
Touch-triggered withdrawal reflexes modeled on the basis of the results of experiments performed with human subjects have been implemented on the NAO robot and validated in the context of safety~\cite{dahl2011}.

Precise control of interaction forces is another important application of tactile systems. In fact conventional approaches using F/T sensors partially fail because they assume that contact happens at a fixed, known location (typically the end-effector). For these reasons several work attempted to integrate contact information from the tactile system and F/T sensors located on the kinematic chain~\cite{iwata02,delprete12,fumagalli12}.
The work in~\cite{delprete12} discusses in details the benefits of an accurate measure of the contact point in the context of force control.

Work on whole-body force control strongly motivates the need for distributed tactile systems. It also highlights the need for such systems to provide accurate contact location in the robot's reference system. For this reason several techniques for automatic calibration of tactile systems have been proposed in the literature. Some of these approaches have been illustrated in Section~\ref{sec:calibration}.

\subsection{Manipulation}

Conventional approaches to grasping are based on accurate models of the objects to be manipulated, including shape, mass, surface texture etc. (see~\cite{okamura_overview_2000} for an overview). In unstructured environments, however, these properties may be incomplete or unavailable. In addition, \textit{model based approaches} require accurate models of the robot hand. Although hand modes may be easier to compute, \textit{model-free} grasping strategies
are expected to be more robust and therefore easier to use. Finally, to cope with uncertainty it would be preferable to have tactile based control strategies that allow the robot to adjust online the grip to an object.

To address these problems, researchers have proposed alternative approaches that rely on tactile or force feedback to evaluate grasp stability~\cite{schill_learning_2012,hang_hierarchical_2016} and to learn control policies that allow the robot to improve the stability of the grasp using feedback control~\cite{sauser_iterative_2012,li_learning_2014}.

In~\cite{schill_learning_2012} the authors trained a SVM to predict whether a grasp is stable or not. The SVM uses features that integrate the current hand configuration (as measured by the encoders on the joints) and tactile feedback. As tactile features the authors used
statistical
moments computed from the tactile image acquired by the sensors on the robot ARMAR III~\cite{armar-tactile-sensors}. A similar approach is adopted in~\cite{hang_hierarchical_2016}, in which Grasp stability is determined using a Gaussian Mixture Model trained off-line as a function of the hand posture and raw tactile data. In these approaches the robot use the learned model to determine, on-line, if a grasp is stable, and apply corrective actions otherwise. Corrective strategies can be implemented using optimization~\cite{hang_hierarchical_2016} or learned by relying on teaching by demonstration~\cite{tegin_demonstration-based_2009, sauser_iterative_2012,regoli2016}, or reinforcement learning~\cite{hoof_learning_2015}.

One of the limitations of approaches based on tactile sensors is that they require accurate detection of contacts. Due to delays in the control loops or insufficient sensitivity the robot may have problems controlling the interaction with the object. Unreliable contact detection may cause the objects to tip over with failures that are difficult to recover. To overcome these problems, alternative approaches based on proximity sensors have been proposed in the literature~\cite{petryk_dynamic_1996,hsiao_reactive_2009}.

\section{Perception and Cognition}
\label{sec:perception}

Conventionally, robotics rely on vision to perceive and identify objects. Although computer vision has recently made remarkable progress, touch can still provide complementary information. 
This is because some material and object properties are simply not accessible from vision (like the object weight), or may be hidden by occlusions. In addition, many properties like surface roughness, 
and even edges and corners, may be confused due to noise or ambiguous, pictorial cues. However, the extraction of tactile features is an intrinsically active process which requires the implementation of appropriate explorative strategies which range from simple behaviors, involving an individual finger (like static contact, contour follow or sliding) to more complex manipulation operations performed with the whole hand.

The definition of \textit{tactile features} is not so straightforward as it is in the case of vision. Several types of features have been proposed in the literature for material and object discrimination. The work of~\cite{hoelscher_evaluation_2015} reports a comprehensive investigation and examination of tactile features. Temperature~\cite{hoelscher_evaluation_2015} and temperature variation~\cite{xu_tactile_2013, hoelscher_evaluation_2015,chu_using_2013} were used with static contact to estimate the type of material by looking at how quickly temperature flowed from the sensor to the contact surface. Static pressure coupled with joint motion provided an estimation of the material softness or compliancy~\cite{xu_tactile_2013, hoelscher_evaluation_2015}. Texture was usually estimated by sliding the finger on the surface and using features that encoded the frequency content of the resulting pressure variation, like variance or features that characterize the frequency spectrum of the signal~\cite{hoelscher_evaluation_2015, xu_tactile_2013, chu_using_2013,jamali_material_2010,sinapov_vibrotactile_2011}. Recently, \cite{baishya_robust_2016} applied deep convolutional neural network to the problem of material discrimination using touch demonstrating that the learned features outperform features based on Fourier analysis for this task.

An important consideration is that texture discrimination requires that the interaction between the sensor and the object generates mechanical vibrations and that the sensor itself can capture such vibrations with sufficient accuracy. This can be facilitated by adding ridges to the surface oft the sensor~\cite{fishel08} and it requires embedding in the sensors transducers that are capable of measuring vibrations like microphones~\cite{fishel08} or accelerometers~\cite{sinapov_vibrotactile_2011}. 

Object discrimination based on material identification has been achieved with high accuracy (\cite{hoelscher_evaluation_2015} reports $97.6\%$ accuracy for a set of 49 objects). The main limitation of the work in the literature is that the interaction between the robot and the object happen in quite controlled situations, with behaviors that are stereotyped and often open-loop. It is still an open challenge how to integrate such techniques with autonomous object manipulation strategies.

\section{Conclusions}
\label{sec:conclusions}

In this Chapter we have provided an overview of the technologies that have been developed in the past years to endow robots with the sense of touch, and how this has contributed to the improvement of robot motor and perceptual skills.

The sense of touch in humanoid robots has gone a long way. Affordable and robust distributed systems of tactile sensors have been successfully built and deployed to cover large areas of various humanoid robots. These systems provide sufficient spatial resolution and are sensitive enough to support sophisticated whole-body behaviors, controlling interaction forces and avoid obstacles. Miniaturized multimodal sensors have been integrated in anthropomorphic hands and have greatly improved the capability of robots to manipulate objects and to perform cognitive tasks like object recognition and material classification. 

Yet, it is clear that to reach human-level performance there are still big challenges to solve which involve all aspects of tactile sensing. Sensors need to become more resilient to endure mechanical stress produced by continuous operation. Perhaps the most difficult challenge is achieving full coverage of the robot body, including moving areas. At this aim new materials and manufacturing techniques for electronic devices will be required to produce stretchable sensors, embedding wiring and electronics. We expect that robotic hands will benefit the most from such technology, with consequent improvement in manipulation capabilities.

Large scale tactile systems also pose non-trivial challenge in terms of power consumption and networking. Solving these problem requires advances in electronic engineering and perhaps new encoding mechanisms to reduce the energy needed to power individual sensors, acquire and route tactile data to the processing units.

%
%
%

%
%
\bibliographystyle{svmult/styles/spmpsci}
\bibliography{main}

\end{document}